\documentclass[10pt,final,journal]{IEEEtran} 
\ifCLASSOPTIONcompsoc
\usepackage[nocompress]{cite}
\else
\usepackage{cite}
\fi
\ifCLASSINFOpdf
\usepackage[pdftex]{graphicx}
\else
\usepackage[dvips]{graphicx}
\fi
\usepackage{amsmath}
\usepackage{algorithmic}

\usepackage{array}
\usepackage{url}

\usepackage[table]{xcolor}
\usepackage{amssymb}
\usepackage[ngerman, english]{babel}
\usepackage[T1]{fontenc}
\usepackage{textcomp}
\usepackage{tikz}
\usepackage{pgfplots}
\pgfplotsset{compat=1.14}
\usepackage{tabularx}
\newcolumntype{Y}{>{\centering\arraybackslash}X}

\newenvironment{leftbar}[1][\hsize]
{%
	\MakeFramed{\hsize{#1}\advance\hsize-\width\FrameRestore}%
}
{\endMakeFramed}

\begin{document}

%
\title{Robust Seed Mask Generation for Interactive\\ Image Segmentation}
%
%
%
%

\author{%
	Mario~Amrehn,
	Stefan~Steidl,
	Markus~Kowarschik,
	Andreas~Maier
	\IEEEcompsocitemizethanks{\IEEEcompsocthanksitem Mario Amrehn, Stefan Steidl, and Andreas Maier are with the Pattern Recognition Lab., Computer Science Department, Friedrich-Alexander University Erlangen-N{\"u}rnberg ({FAU}), Germany%
		\IEEEcompsocthanksitem Markus Kowarschik is with Siemens Healthcare GmbH, Forchheim, Germany
		\IEEEcompsocthanksitem Andreas Maier is with the Erlangen Graduate School in Advanced Optical Technologies (SAOT), Germany
	\IEEEcompsocthanksitem E-mail: see https://www5.cs.fau.de/{\textasciitilde}amrehn}%
}

%
%

\markboth{2017 IEEE Nuclear Science Symposium and Medical Imaging Conference}%
{Amrehn \MakeLowercase{\textit{et al.}}: 
Robust Seed Template Generation for Interactive Image Segmentation}
\twocolumn[
\begin{@twocolumnfalse}
	\maketitle
\begin{abstract}	
\noindent In interactive medical image segmentation, anatomical structures are extracted from reconstructed volumetric images.
The first iterations of user interaction traditionally consist of drawing pictorial hints as an initial estimate of the object to extract. 
Only after this time consuming first phase, the efficient selective refinement of current segmentation results begins. 
Erroneously labeled seeds, especially near the border of the object, are challenging to detect and replace for a human and may substantially impact the overall segmentation quality. 
We propose an automatic seeding pipeline as well as a configuration based on saliency recognition,
in order to skip the time-consuming initial interaction phase during segmentation.
A median Dice score of $68.22\,\%$ is reached before the first user interaction on the test data set with an error rate in seeding of only $0.088\,\%$.
\end{abstract}
\end{@twocolumnfalse}
]
%


%

{
	\renewcommand{\thefootnote}{\hspace{-0.28265cm}\fnsymbol{footnote}}
	\footnotetext[1]{Mario Amrehn, Stefan Steidl, and Andreas Maier are with the Pattern Recognition Lab., Computer Science Department, Friedrich-Alexander University Erlangen-N{\"u}rnberg ({FAU}), Germany}
	\footnotetext[1]{Markus Kowarschik is with Siemens Healthcare GmbH, Forchheim, Germany}
	\footnotetext[1]{Andreas Maier is with the Erlangen Graduate School in Advanced Optical Technologies (SAOT), Germany}
	\footnotetext[1]{E-mail: see https://www5.cs.fau.de/{\textasciitilde}amrehn}
}

\section{Introduction}\label{sec:introduction} 
%
\IEEEPARstart{S}{egmentation} is a fundamental part of semantic image analysis.
During segmentation, each image element is mapped to one of $N$ pre-defined class labels.
In medical image processing, 
two-class problems ($N:=2$) are upon the most common use-cases for image segmentation.
Here, significant objects, often coherent anatomical structures like liver tumors ({HCC}), are extracted from image background based on local image features.
Interactive image segmentation provides a good trade-off between the accuracy of manual segmentation and the speed and scalability of automatic segmentation techniques.
In a cooperative human computer interaction ({HCI}) workflow, 
the current segmentation result's similarity to the desired object boundary is improved iteratively. 
The user incrementally adds pictorial hints to the seed mask, 
which, in addition to the volumetric image, is utilized as the input for the segmentation technique.
After each additional hint, the computation of the next segmentation result is started, until a user-defined stopping criterion is reached.

Popular workflows like \cite{rother2004grabcut, vezhnevets2005growcut, grady2006random} start the interactive process 
without any prior seed mask.
Several usability studies \cite{jain2013predicting, andrade2015supervised, amrehn2018usability} illustrate, that users spend significant amounts of time and effort in the beginning of the interactive workflow to achieve segmentation accuracies which would also be reachable by fully automated systems \cite{militzer2010automatic}.
Only after this first phase, an interactive approach can outperform other segmentation techniques. 
Utilizing standard automated segmentation for a generic seed selection step prior to user defined seeding is challenging, since most seeded segmentation techniques require a truthful labeling of all input seed points.
Given invalid labels, the ground truth segmentation may not be reachable even if the user is not restricted on time or number of interactions for the further process.
An ideal pre-seeding technique therefore, (1) automates the task of initial seed placement, maximizing the initial segmentation's quality, while (2) minimizing the error of falsely defined object labels.
Such seeding methods, embedded in a common pipeline framework, are proposed and evaluated in this paper.

\section{Materials and Methods}
\begin{figure}[t]
	\includegraphics[width=\columnwidth]{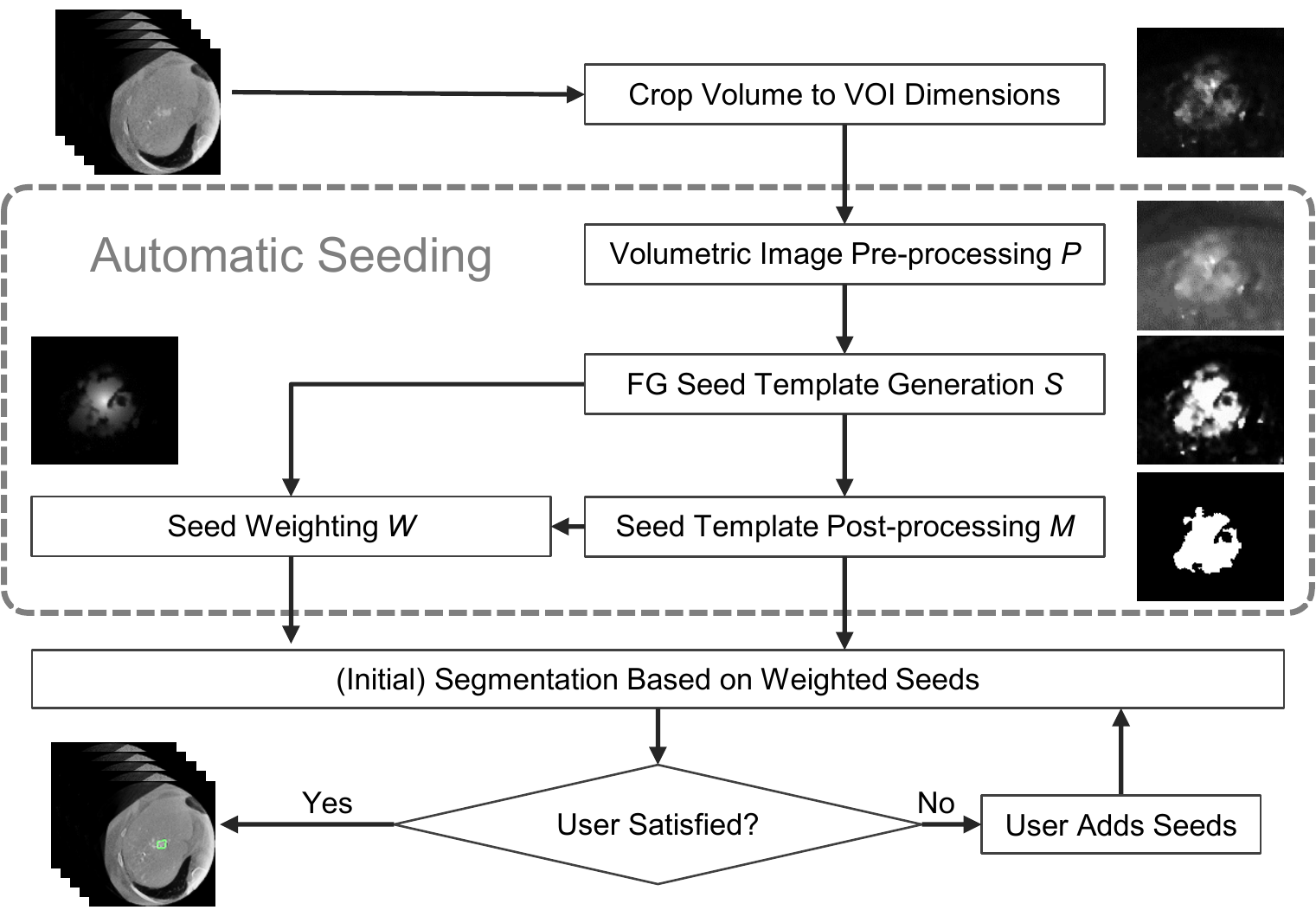}
	\caption{Interactive segmentation workflow.
		The dashed rectangle indicates the automated seeding process defined by \mbox{$(\mathbf{P},\mathbf{S},\mathbf{W},\mathbf{M})$}.}
	\label{fig:interactive_seg_flowchart}
\end{figure}

The proposed workflow for automated seeding consists of four procedural steps \mbox{$(\mathbf{P},\mathbf{S},\mathbf{W},\mathbf{M})$}, as depicted in Fig. \ref{fig:interactive_seg_flowchart}.
Image pre-processing $\mathbf{P}$ is performed by bilateral filtering.
Seeding $\mathbf{S}$ is performed separately for foreground ({FG}) seeds and background ({BG}) seeds.
Given an object which is in its entirety inside the image volume, 
valid {BG} seeds can simply be generated along the border of the volume.
For {FG} seeding, we focus on three types of methods: Otsu thresholding $\mathbf{S}_o$ \cite{otsu1979threshold}, Gaussian mixture model ({GMM}) based segmentation $\mathbf{S}_g$, and binarized saliency detection $\mathbf{S}_{\{r,t,m,f\}}$.

\begin{figure*}[t]
	\resizebox{\textwidth}{!}{%
		\begin{tabular}{cc}
			\includegraphics[width=\columnwidth]{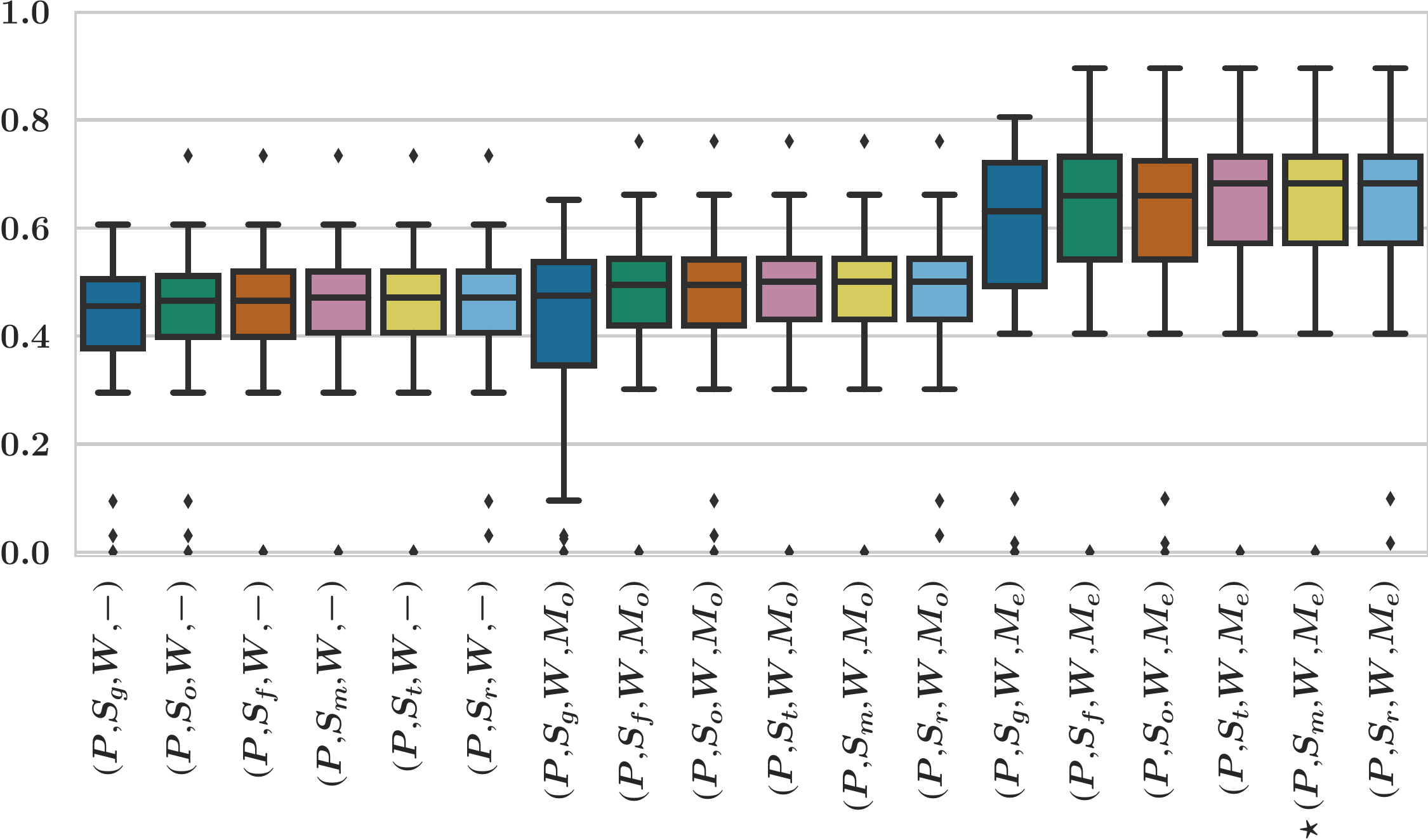} &
			\includegraphics[width=\columnwidth]{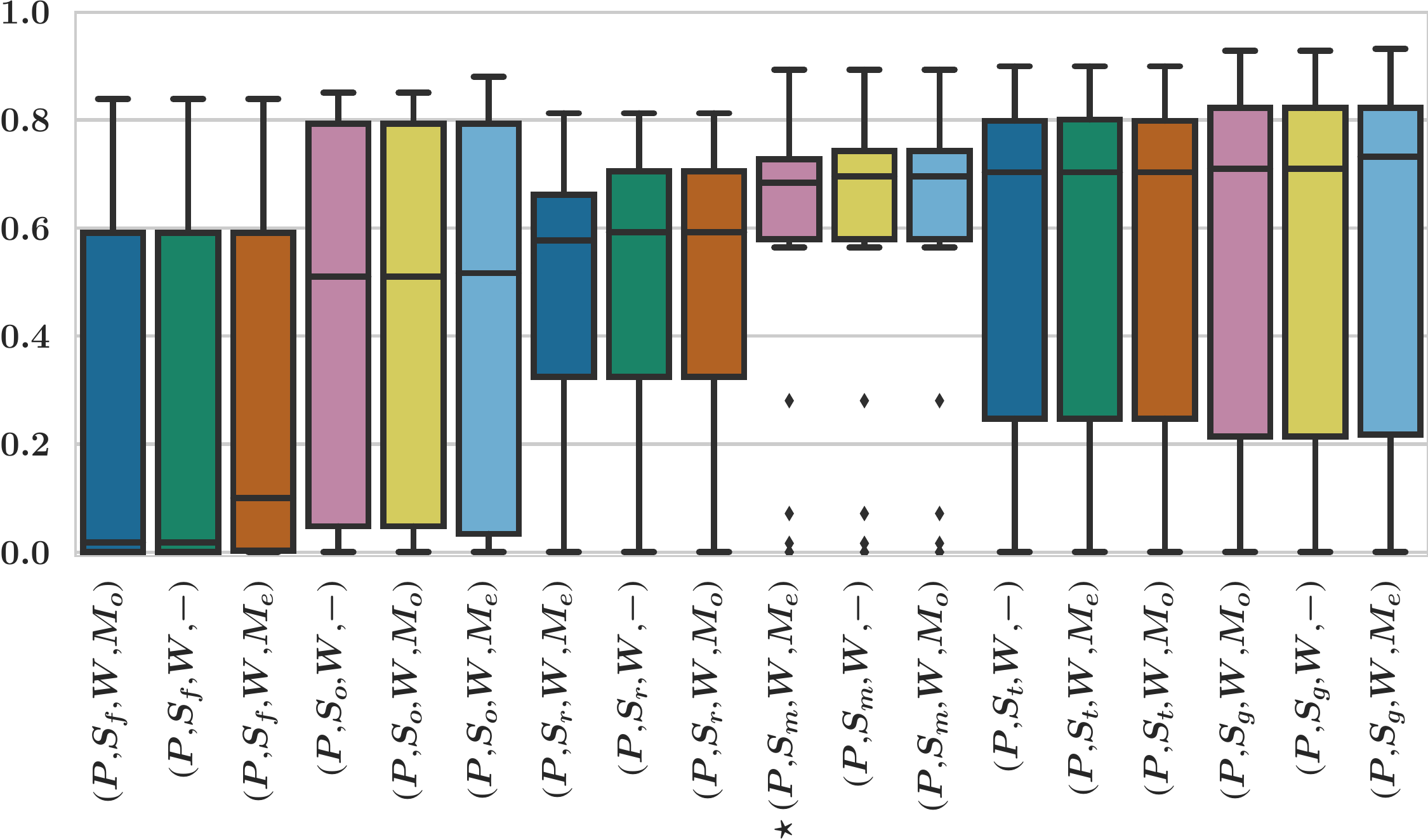}
		\end{tabular}%
	}%
	\caption{Evaluation of {GC} (left) and {RW} (right) segmentations' Dice scores.
		Proposed method \mbox{($\mathbf{P}$, $\mathbf{S}_m$, $\mathbf{W}$, $\mathbf{M}_e$)} is highlighted by $\star$.}%
	\label{fig:dice_scores}%
\end{figure*}

For $\mathbf{S}_g$, a {GMM} is trained on the intensity values of the current image.
Subsequently, the Gaussian best describing {FG} seeds is select by comparing the median intensities of image elements which associated probability density function ({PDF}) value is above a certain threshold.
The Gaussian distribution with maximal median value is chosen for {FG} seed generation.
The mean of the maximum and median values of the {PDF} are then utilized as threshold for a binarization of the probability map.

The goal of saliency detection is to emphasize and outline the largest salient object in the volumetric image, 
while disregarding high frequencies which may arise from noise and texture artifacts.
Four popular saliency methods are compared:
$\mathbf{S}_r$ \cite{zhu2014saliency}, based on connectivity quantization as a boundary prior for robust background detection, 
frequency-tuned detection $\mathbf{S}_t$ \cite{achanta2009frequency}, 
a minimum barrier detection $\mathbf{S}_m$ \cite{zhang2015minimum}, and 
saliency filters $\mathbf{S}_f$ \cite{perazzi2012saliency} utilizing low-level image features like contrast. 
Binarization is performed by Otsu thresholding of the $10\,\%$ largest saliency scores per image.

Seeds are weighted $\mathbf{W}$ by a Gaussian kernel, shifted to the center of mass of the current seed mask. 
Weights for non-seed image elements are set to zero. 
Morphological post-processing is conducted by binary opening $\mathbf{M}_o$, binary erosion $\mathbf{M}_e$, or omitted. 
For the evaluation, a segmentation is performed on the seeds via RandomWalker ({RW}) \cite{grady2006random} and GrowCut ({GC}) \cite{vezhnevets2005growcut, amrehn2016comparative}.
For {GC}, the weighted seed mask is utilized as the strength map. 
An evaluation is conducted utilizing $28$ fully annotated volumetric {HCC} data sets.

\section{Results and Discussion}
\begin{figure}
	\resizebox{\columnwidth}{!}{
		\includegraphics{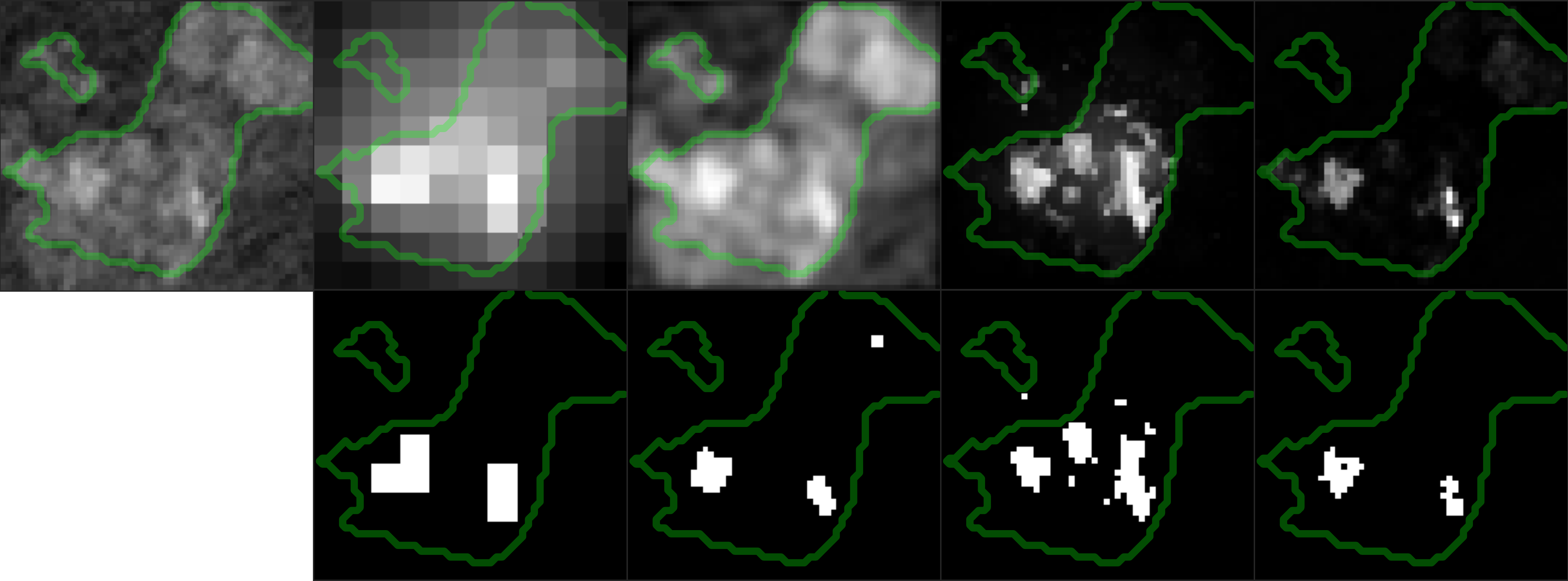}
	}
	\begin{tabularx}{\columnwidth}{YYYYY}
		& $\mathbf{S}_r$ \cite{zhu2014saliency} & $\mathbf{S}_t$ \cite{achanta2009frequency} & $\mathbf{S}_m$ \cite{zhang2015minimum} & $\mathbf{S}_f$ \cite{perazzi2012saliency} %
	\end{tabularx}%
	\caption{Saliency maps (upper row) from input image (upper left) utilizing different detection techniques $\mathbf{S}_{\{r,t,m,f\}}$.
		Seed masks (lower row) are obtained via thresholding.
		Ground truth object contour line depicted in green.}%
	\label{fig:saliency_phase_qual_results}%
\end{figure}

\begin{figure}
	\resizebox{\columnwidth}{!}{%
		\includegraphics{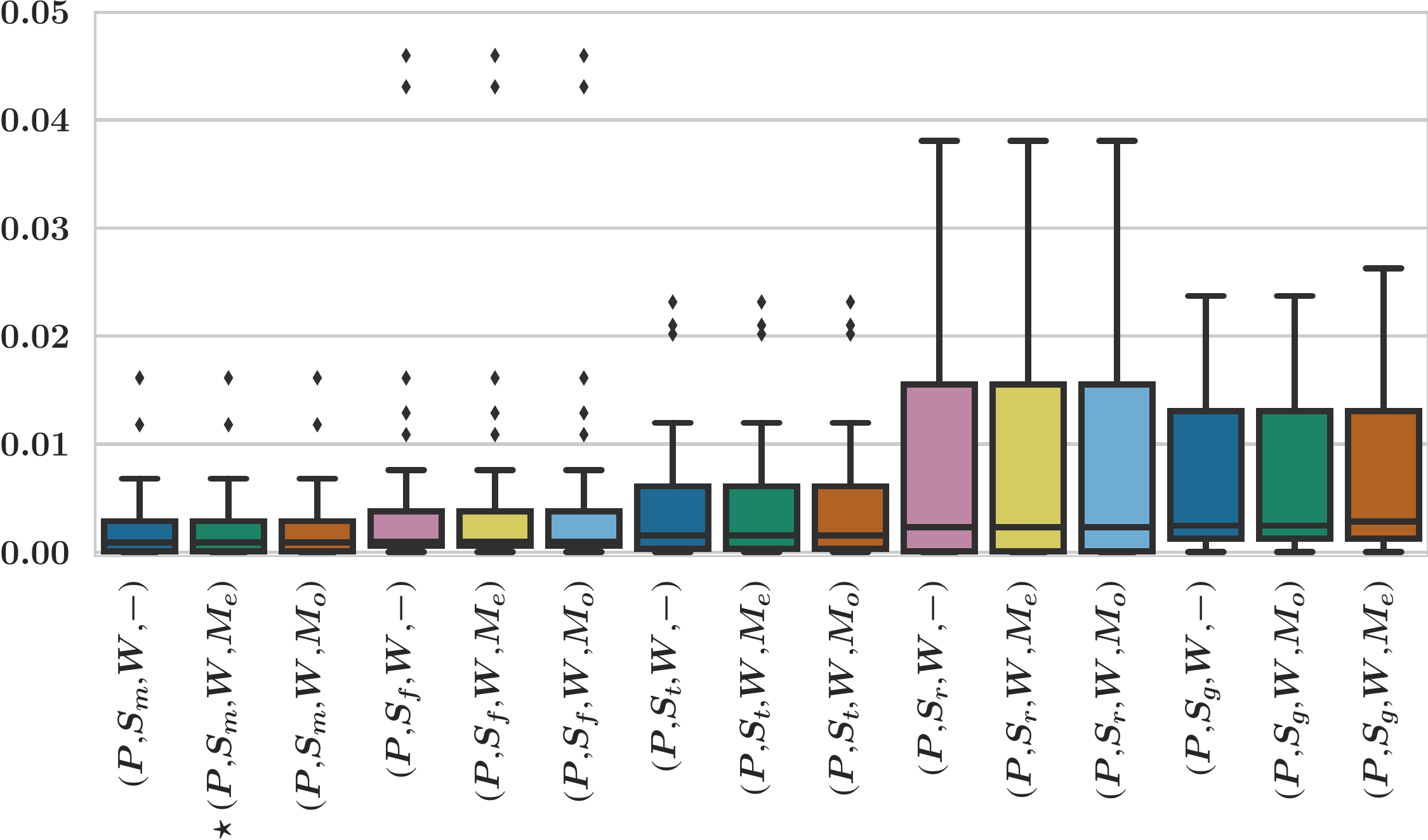}%
	}%
	\caption{False negative rate ({FNR}) of object seed points. %
		Mean {FNR} medians per method: $0.09\,\%$ ($\mathbf{S}_m$), $0.10\,\%$ ($\mathbf{S}_f$), $0.16\,\%$ ($\mathbf{S}_t$), $0.23\,\%$ ($\mathbf{S}_r$), $0.30\,\%$ ($\mathbf{S}_g$).}%
	\label{fig:fpr_seeding_score}%
\end{figure}

Qualitative results of the saliency binarization phase are depicted in Fig.\ \ref{fig:saliency_phase_qual_results}.
Here, only $\mathbf{S}_m$ produces an invalid {FG} seed, which, however, is eliminated during the subsequent post-processing $\mathbf{M}$.
The segmentations via {GC} and {RW}, which are based on the automated pre-seeding, are depicted in Fig.\ \ref{fig:dice_scores}.
Method \mbox{($\mathbf{P}$, $\mathbf{S}_m$, $\mathbf{W}$, $\mathbf{M}_e$)} achieves the highest median Dice score for {GC} segmentation.
\mbox{($\mathbf{P}$, $\mathbf{S}_m$, $\mathbf{W}$, $\mathbf{M}_e$)} median Dice score for {RW} is $93.4\,\%$ of the maximal score reached by \mbox{($\mathbf{P}$, $\mathbf{S}_g$, $\mathbf{W}$, $\mathbf{M}_e$)}, however, yields more robust results due to $26.2\,\%$ reduced standard deviation.
The {FPR} of the generated seeds w.\,r.\,t.\ the ground truth are depicted in Fig.\ \ref{fig:fpr_seeding_score}.

\section{Conclusion}
An automated seeding pipeline was defined and evaluated, which supports various saliency detection based as well as {GMM} and thresholding based methods. 
An extensive comparison of pipeline element selections resulted in the proposition of configuration \mbox{($\mathbf{P}$, $\mathbf{S}_m$ \cite{zhang2015minimum}, $\mathbf{W}$, $\mathbf{M}_e$)} for pipeline usage.

Previously stated goal (1) for automated seeding is reached due to the high quality segmentations yielded by \mbox{($\mathbf{P}$, $\mathbf{S}_m$, $\mathbf{W}$, $\mathbf{M}_e$)}.
The low {FPR} results of \mbox{($\mathbf{P}$, $\mathbf{S}_m$, $\mathbf{W}$, $\mathbf{M}_e$)}, crucial for successful automated seed placement, fulfills the second stated goal (2). 

\newpage



\bibliographystyle{IEEEtran}
\vspace{0.080909cm}
Disclaimer: the concept and software presented in this paper are based on research and are not commercially available.
Due to regulatory reasons its future availability cannot be guaranteed.
%
\appendices
\section{}
%
\begin{figure}[h]
	\caption{Selected results: saliency maps (upper rows) from input image (leftmost column) utilizing different detection techniques $\mathbf{S}_{\{r,t,m,f\}}$. %
		Seed masks (lower rows) are obtained via thresholding and weighting. 
		The annotated contour lines of the ground truth segmentation are depicted in green.\vspace{10pt}}%
	\resizebox{\columnwidth}{!}{%
		\begin{tabular}{l}
			\includegraphics{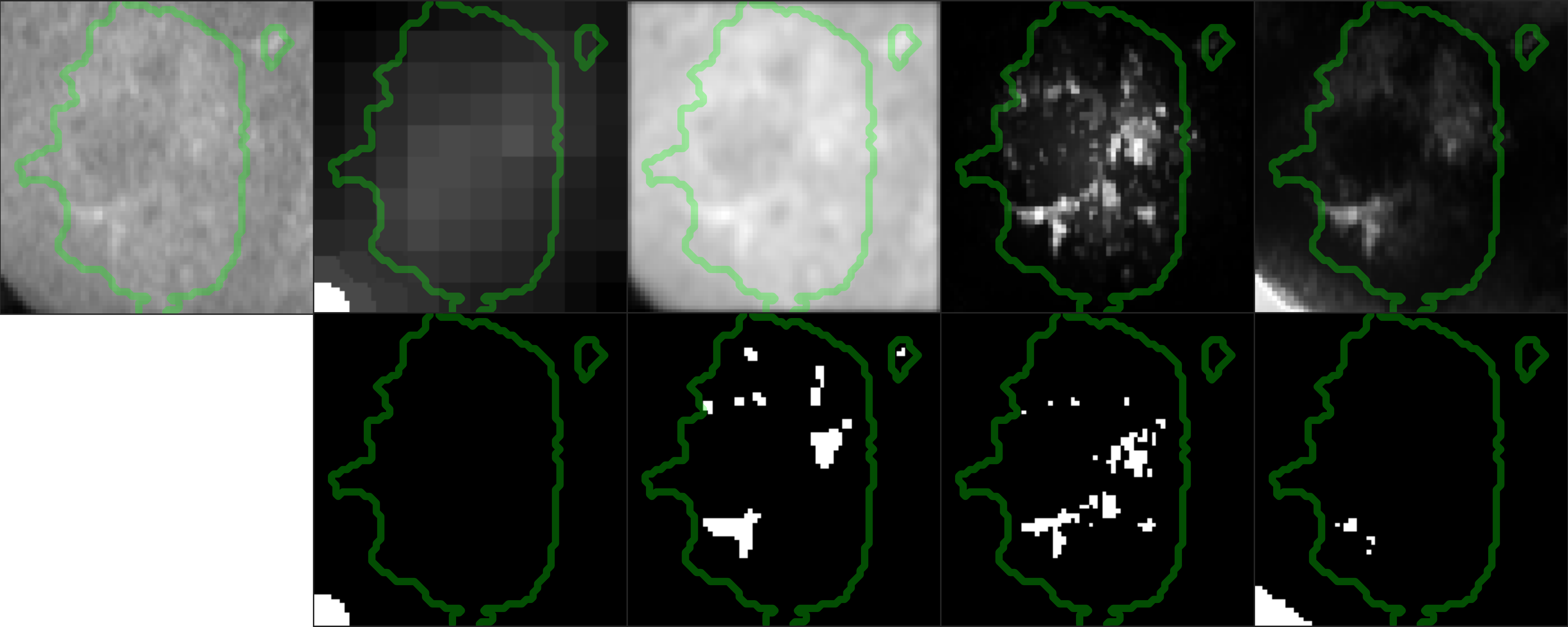}\\
			\includegraphics{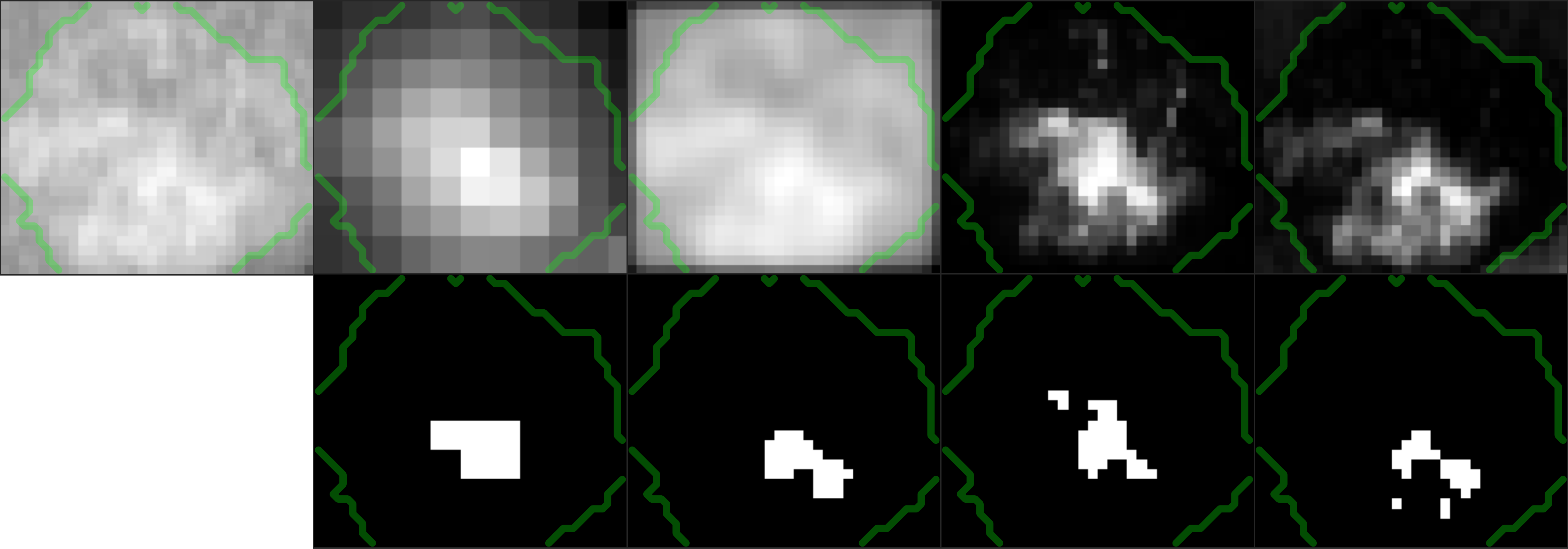}\\
		\end{tabular}
	}
	\begin{tabularx}{\columnwidth}{YYYYY}
		& $\mathbf{S}_r$ \cite{zhu2014saliency} & $\mathbf{S}_t$ \cite{achanta2009frequency} & $\mathbf{S}_m$ \cite{zhang2015minimum} & $\mathbf{S}_f$ \cite{perazzi2012saliency} %
	\end{tabularx}%
\end{figure}
\begin{figure}
	\resizebox{\columnwidth}{!}{%
		\begin{tabular}{l}
			\includegraphics{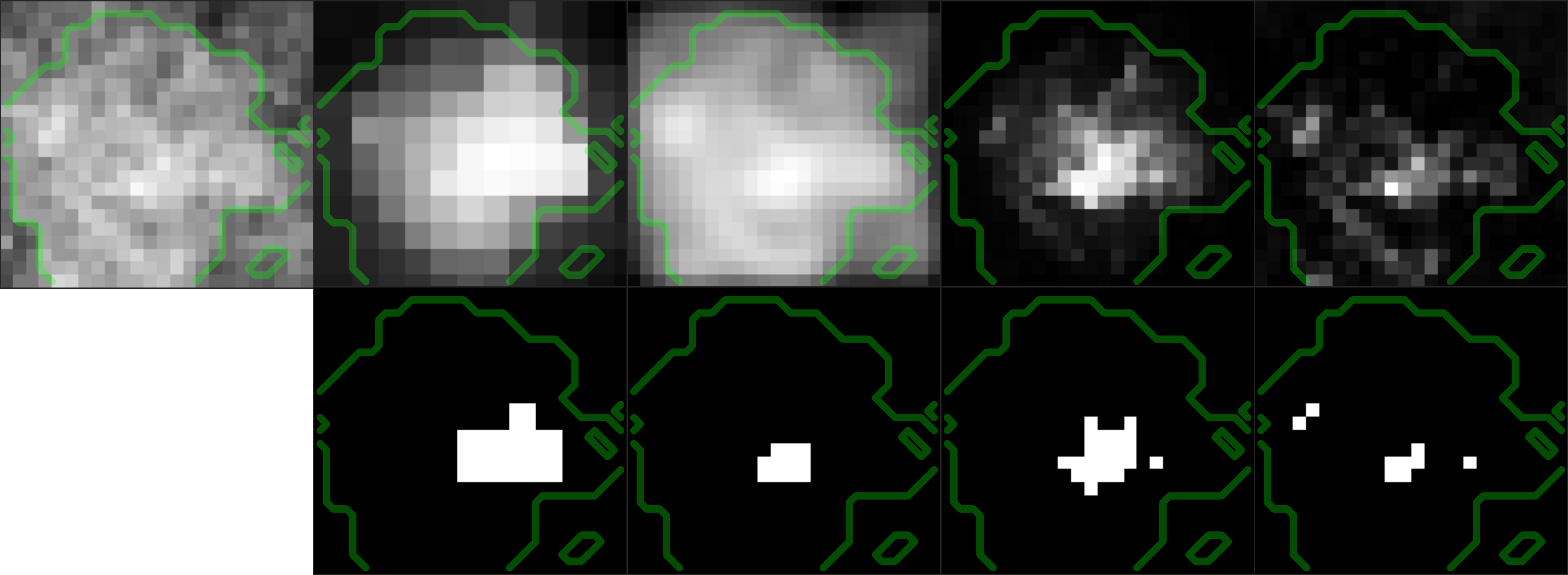}\\
			\includegraphics{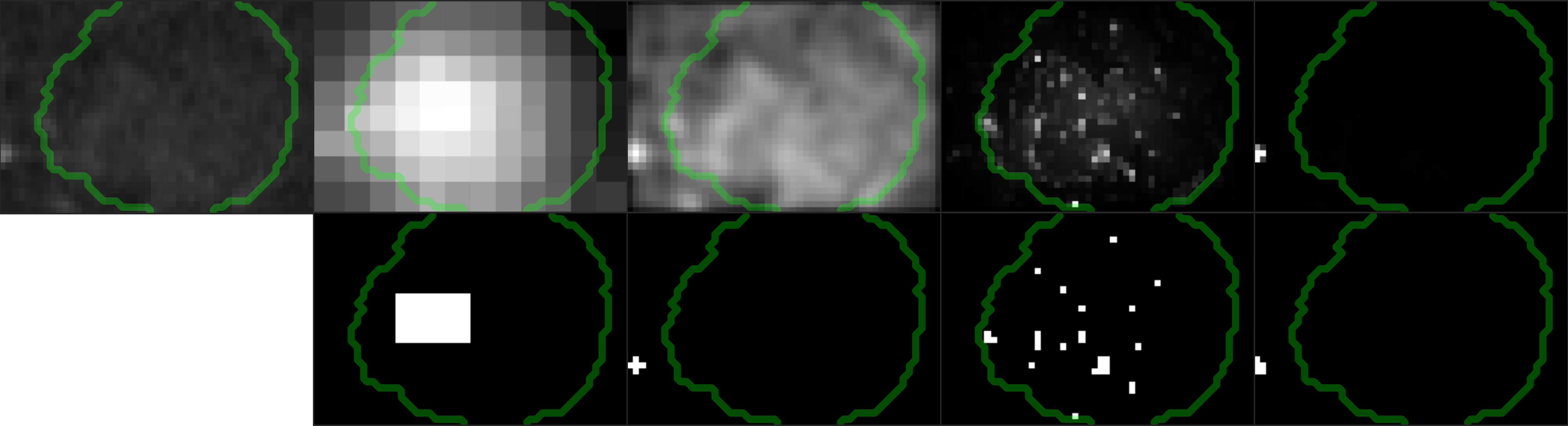}\\
			\includegraphics{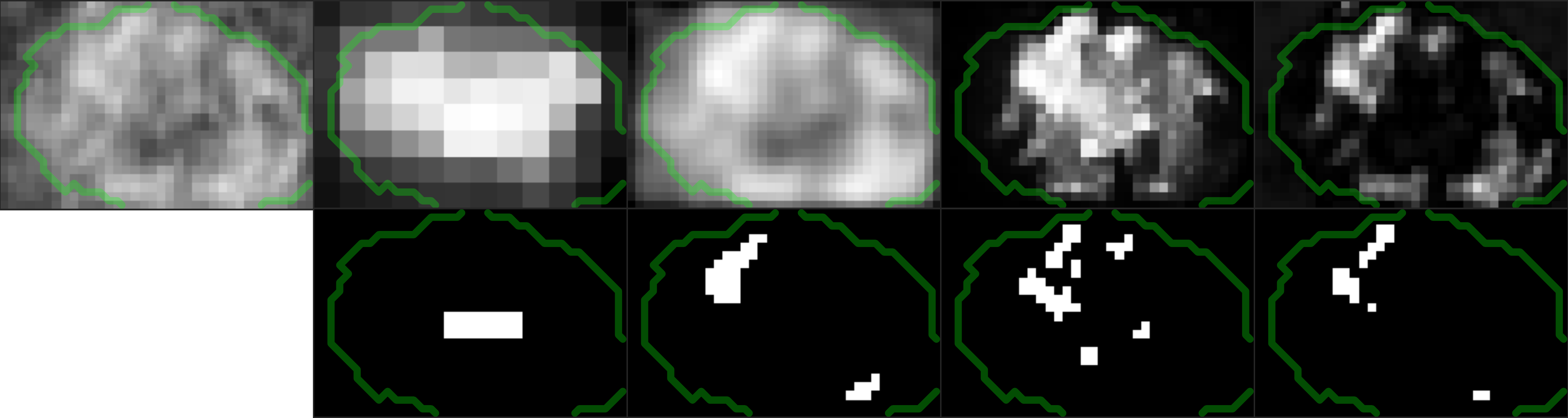}\\
			\includegraphics{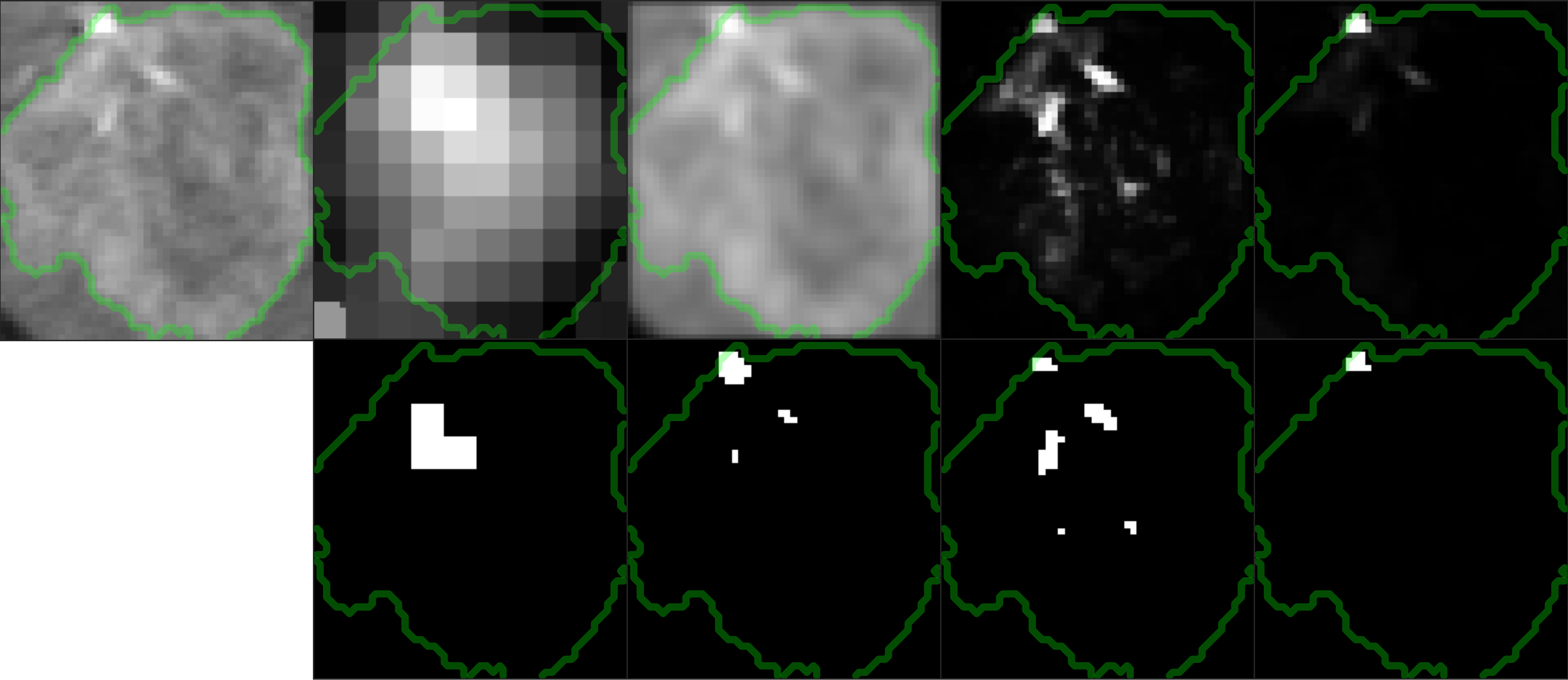}\\
			\includegraphics{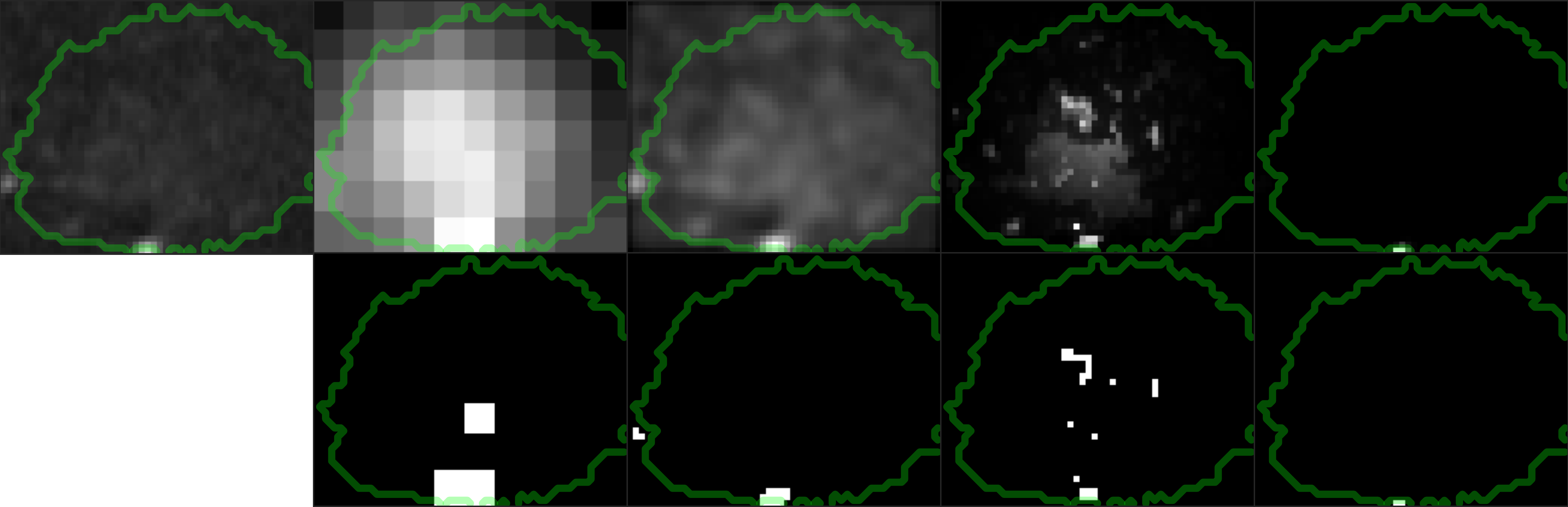}\\
			\includegraphics{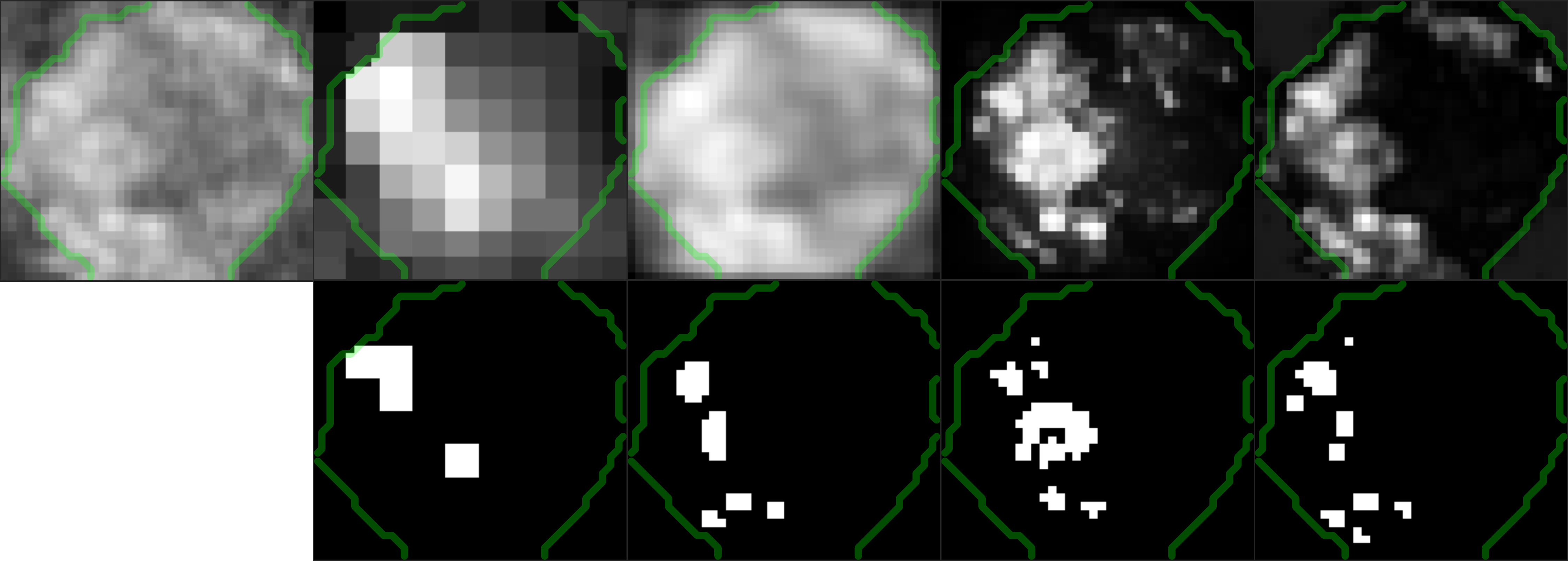}\\
			\includegraphics{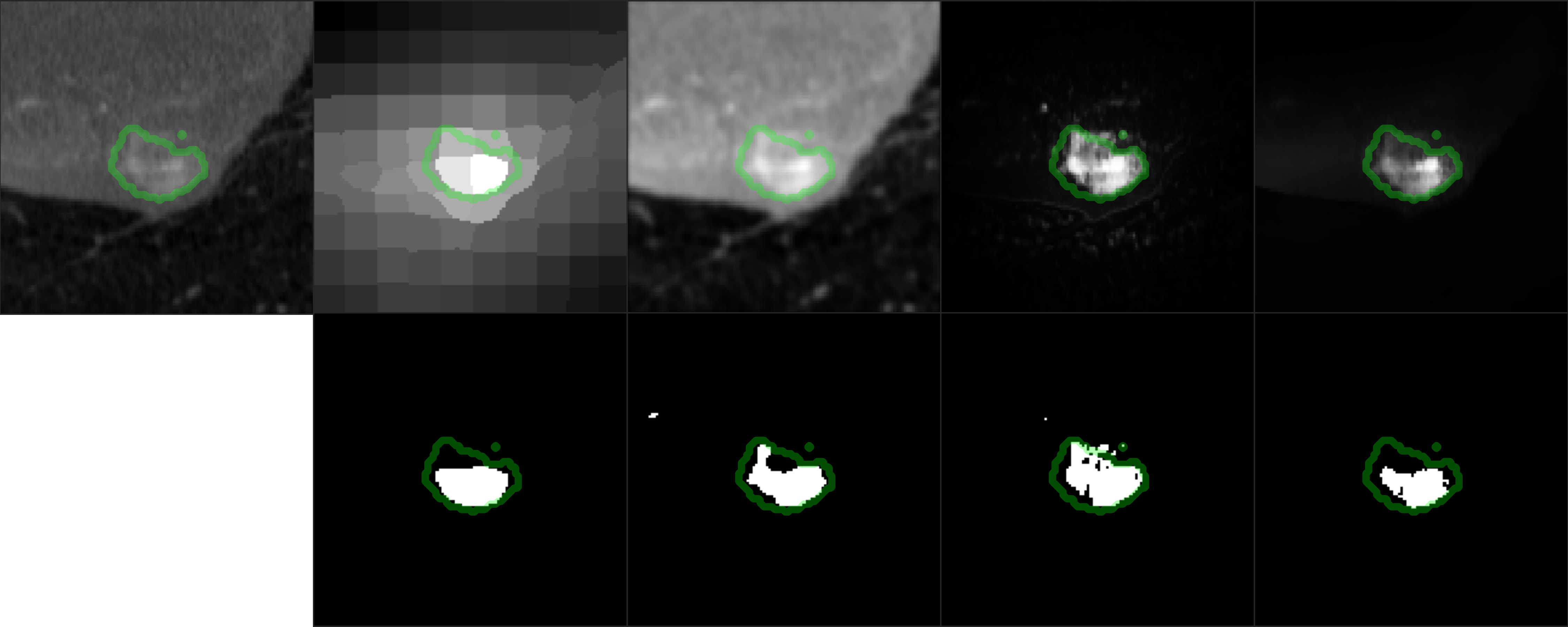}\\
			\includegraphics{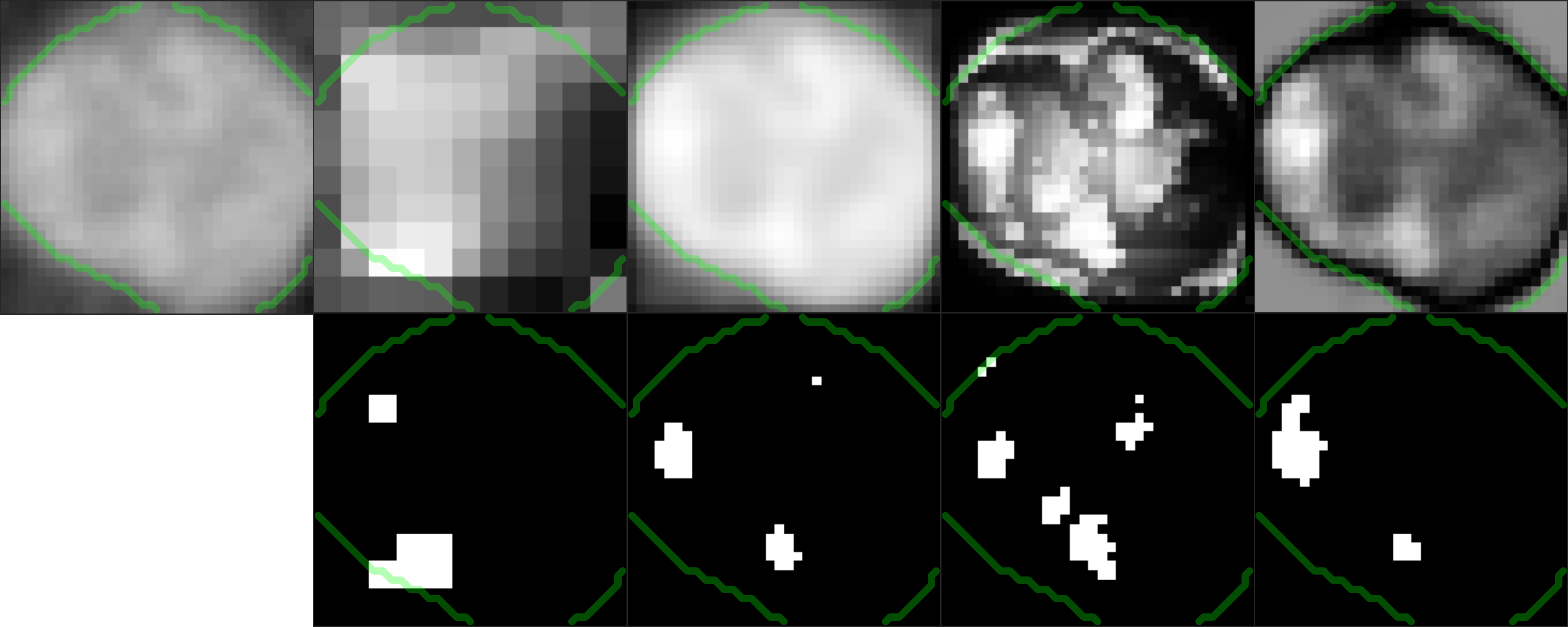}\\
		\end{tabular}
	}
	\begin{tabularx}{\columnwidth}{YYYYY}
		& $\mathbf{S}_r$ \cite{zhu2014saliency} & $\mathbf{S}_t$ \cite{achanta2009frequency} & $\mathbf{S}_m$ \cite{zhang2015minimum} & $\mathbf{S}_f$ \cite{perazzi2012saliency} %
	\end{tabularx}%
\end{figure}

\end{document}